\documentclass[letterpaper, 10 pt, conference]{ieeeconf}  % Comment this line out if you need a4paper

\IEEEoverridecommandlockouts                              % This command is only needed if 
\usepackage{hyperref}
\usepackage{graphicx}
\usepackage{amsmath}
\usepackage{amssymb}
\usepackage{color}
\usepackage{parskip}
\usepackage{mwe}
\usepackage[utf8]{inputenc}

\usepackage{tikz}
\usepackage{pgfplotstable}
\usepackage{makecell}
\pgfplotsset{compat=1.14}
\usepgfplotslibrary{statistics}
\usepgfplotslibrary{groupplots}

\usepackage{color}

\title{\LARGE \bf
Improved Visual Localization via Graph Smoothing
}

\author{Carlos Lassance$^{1,2}$, Yasir Latif$^{3}$, Ravi Garg$^{3}$, Vincent Gripon$^{1,2}$ and Ian Reid$^{3}$% <-this % stops a space
\thanks{*This work was supported by FASIC, the Britanny region. Computations were performed using Nvidia GPUs, courtesy of Nvidia.}% <-this % stops a space
\thanks{$^{1}${\tt\small IMT Atlantique and Lab-STICC, Brest, France}}%
\thanks{$^{2}${\tt\small Université de Montréal and Mila, Canada}}%
\thanks{$^{3}${\tt\small University of Adelaide, Australia}}%
}

\begin{document}

\maketitle
\thispagestyle{empty}
\pagestyle{empty}

%%%%%%%%%%%%%%%%%%%%%%%%%%%%%%%%%%%%%%%%%%%%%%%%%%%%%%%%%%%%%%%%%%%%%%%%%%%%%%%%
\begin{abstract}
%Vision based localization is the problem of inferring the pose of the camera given a single image. This is generally addressed using regression deep learning models pretrained on reference images with known poses. Another approach consists in comparing the query image against the reference image database. The pose of the query image is then estimated from those of the set of retrieved matches. The latter approach assumes that images showing the same objects, and thus coming from similar places, would exhibit similar feature vectors. In this work, we introduce a method to enhance the performance of this process by taking into account additional information including GPS coordinates, consecutiveness in the acquisition process and descriptor similarity of pairs of images in the reference or query database. The method consists in constructing a graph based on this information, that is used to smooth the feature vectors of reference and/or query images. Using two large scale experiments, we show that the proposed method is able to significantly improve the accuracy of the process.

Vision based localization is the problem of inferring the pose of the camera given a single image. One solution to this problem is to learn a deep neural network to infer the pose of a query image after learning on a dataset of images with known poses. Another more commonly used approach rely on image retrieval where the query image is compared against the database of images and its pose is inferred with the help of the retrieved images. The latter approach assumes that images taken from the same places consists of the same landmarks and, thus would have similar feature representations. These representation can be learned using full supervision to be robust to different variations in capture conditions like time of the day and weather. In this work, we introduce a framework to enhance the performance of these retrieval based localization methods by taking into account the additional information including GPS coordinates and temporal neighbourhood of the images provided by the acquisition process in addition to the descriptor similarity of pairs of images in the reference or query database which is used traditionally for localization. Our method constructs a graph based on this additional information and use it for robust retrieval by smoothing the feature representation of reference and/or query images. We show that the proposed method is able to significantly improve the localization accuracy on two large scale datasets over the baselines.
\end{abstract}

%%%%%%%%%%%%%%%%%%%%%%%%%%%%%%%%%%%%%%%%%%%%%%%%%%%%%%%%%%%%%%%%%%%%%%%%%%%%%%%%
\section{Introduction}

%Vision-Based Localization (VBL)~\cite{PIASCO201890} is the problem of retrieving the pose of the camera (position and/or orientation) given an input (query) image.
%VBL can be used for indoor localization, improving accuracy of vehicle tracking, or visual maps creation. The approaches for addressing VBL can be broadly divided into two categories~\cite{PIASCO201890}:
%Various methods have been proposed in the literature, which can be divided in two categories:
%\begin{enumerate}
%    \item Direct methods: the pose information is retrieved directly from the visual query. This can be thought of as a regression problem and has been addressed using deep learning techniques. A set images with known poses at training time is used to learn a mapping from raw pixel to poses.
%    \item Indirect methods: the pose information is inferred from the visual query using a reference database, where each image in the database has an associated pose. This can be seen as an image retrieval problem as the main task is to find images that come from the same location. Once a match or set of matches is found, the pose for the current query is computed as a function of the poses of  the retrieved images.
%\end{enumerate}

Vision-Based Localization (VBL)~\cite{PIASCO201890} is the problem of retrieving the location and orientation (pose) of the camera which will generate a given query image. VBL can be used to improve accuracy of vehicle tracking as well as for accurate visual maps creation via loop closure. The approaches for addressing VBL can be broadly divided into two categories~\cite{PIASCO201890}:
\begin{enumerate}
    \item Direct methods: These methods directly retrieve pose from the visual query -- usually by solving a regression problem. Use of deep learning techniques replacing this regressing have become prevalent forming the current state of the art where a set images with known poses are used to learn a mapping from raw pixel colors to image poses.
    \item Indirect methods: In these methods, the pose information is inferred from the visual query using a reference database, where each image in the database has an associated pose. This can be seen as an image retrieval problem where the aim is to find images in the support set that might have been taken from the same location as that of the query image. Once a match or set of matches is found, the pose for the query image is computed as a function of the poses of the retrieved images.
\end{enumerate}

%Vision-Based Localization has received a lot of attention recently, especially regarding the design of deep neural networks to either perform the regression (direct)~\cite{brahmbhatt2018geometry,kendall2015posenet} or to generate latent representations, which are resilient to appearance changes, for the image retrieval problem~\cite{arandjelovic2016netvlad}. 

Use of the deep learning for VBL approaches has recently received a lot of attention  both in terms of learning direct image to pose mappings~\cite{brahmbhatt2018geometry,kendall2015posenet} or to generate latent representation that are resilient to appearance changes~\cite{arandjelovic2016netvlad}. A major drawback of direct approaches is their inability to generalize to previously unseen locations. Even small difference of query pose from the training set can cause gross localization errors and appending new query locations to the dataset for direct approach will require retraining the whole pose estimation network from scratch. On the contrary, indirect methods generalize well to new data without the need for this retraining. 

However, the retrieval based indirect VBL is challenging because it is extremely difficult to learn a representation which is resilient to a huge amount of appearance variations. Moreover, if the pose of the query image is relatively different from the corresponding images in the database a correct match needs to be found by interpolation in the latent representations with additional information. Sequence to Sequence matching approaches like \cite{seqslam} have been proposed for making the retrieval more robust in these cases, but the success of these approaches simply rely on having a large number of images in the query set. 

Proposing a very robust method for sequence/one to one image matching for localization due to external factors (e.g. different image acquisition conditions) is a hard problem. A principled solution will be capable of smartly interpolating the latent space for retrieval, given some additional information about the data. The large support dataset against which a query can be retrieved generally comes with rich information such as GPS location of the image and temporal order in which the images are captured. 

In this work, we propose an indirect visual localization method that takes advantage of the additional information that might be available for each image in the database, including GPS coordinates, consecutiveness in the acquisition process and similar latent representations.
%benefits from the extra information contained in the database. 
This is particularly interesting for a robotics setting, 
where images are almost always acquired sequentially from a camera mounted on a vehicle. This sequential nature of the acquisition process suggests that images closer in time should also be close in the latent representation. 
Indeed, it is intuitive that temporally adjacent images have similar latent representation. 
Additional information such as GPS coordinates, if available, can aid in encoding global relationships between images in the database.
We show that by considering such relationships between images,
localization accuracy can be increased. 
Moreover, enhancements can be achieved using only minor adjustments to the inference process.
Specifically, we exploit relationships via a graph smoothing operation \cite{shuman2013emerging} on top of pre-learned deep representations extracted from NetVLAD~\cite{arandjelovic2016netvlad}.
The graph smoothing operation is derived from the Graph Signal Processing (GSP) framework~\cite{shuman2013emerging}, and takes advantage of the graph based representation of the problem.
In this graph, each vertex is associated one-to-one with an image (or equivalently its latent representation). Edges model relations between images and are derived from the additional source of information (e.g. temporal adjacency, GPS, similar latent representations).

%where the edges are generated by the additional information available from the database and the node information which is given by the NetVLAD representation.
Interestingly, the proposed method can be seen as a fine-tuning of the representations that does not require additional learning, allowing this operation to be possibly executed on a resource constrained system.
%\YLc{We might want to not claim this. What can be said is that this might be possible to refine on a resource constrained system, as opposed to deploying.} \CLc{Yes this is true. I have rephrased this to better reflect that.}

\textbf{Main contributions:}
The main contributions of this work are two-fold:
\begin{enumerate}
    \item We apply Graph Signal Processing techniques to the problem of indirect visual localization. To the best of our knowledge, we are the first to bring together the area of Graph Signal Processing and Visual Based Localization.
        %We introduce the first GSP-based framework for the task of indirect visual localization. 
    \item Through experiments on real-world datasets, we demonstrate the efficacy of the proposed method in improving the accuracy of the indirect VBL process on large scale datasets.
\end{enumerate}

The rest of the paper is organized as follows:
%The outline is as follows:
we present a brief overview of related techniques in Section~\ref{rw}.
In Section~\ref{methodo}, we formally introduce the proposed method and discuss its properties. 
In Section~\ref{experiments}, we derive and discuss experiments. 
In Section~\ref{conclusion}, we conclude the work and discuss future directions.

\section{Related Work}
\label{rw}

Visual localization is a well studied problem in the vision community and a recent survey can be found in~\cite{PIASCO201890}.
Traditional methods address the problem using point features using a Bag-of-Words (Bow) approach where each image is represented as histogram of visual word occurrences.
Efficient indexing methods then allow retrieving images with similar features and a relate pose computation via the essential matrix. 
However, such methods can be adversely affected by changes in condition such as weather, time of the day and long term changes such as structure of the scene. 

\textbf{Deep learning in direct visual localization:}
with the recent revival of deep learning, work has focused on formulating VBL as an end-to-end learning problem where the pose of the image is regressed from the raw pixels via a deep neural network \cite{kendall2015posenet}.  
Other works such as \cite{brachmann2018learning} have explored scene coordinate regressed followed by RANSAC to compute the camera pose via 3D to 2D correspondences. This has shown great improvement over the end-to-end approach. 

\textbf{Deep learning in indirect visual localization:} as mentioned in the introduction, various methods in the literature focus on deep learning for generating good embeddings for indirect visual localization, such as NetVLAD~\cite{arandjelovic2016netvlad}. In this work, we build on top of these representations, though the proposed method could be adapted to any latent representation of the images. Its main advantage is that it is not required to perform any additional training. Recent work in robotics \cite{doan2019scalable} has shown that using sequence information in Bayesian filtering approach, the accuracy of indirect methods can be vastly improved, even outperforming direct methods.

\textbf{Graphs in visual localization:} works~\cite{torii2011visual,brahmbhatt2018geometry,Cao_2013_CVPR} have used graphs to increase the performance of visual localization methods in various ways. One example is the re-ranking of candidates in indirect VBL, where one can use a graph to perform a ranking that takes into account more than one image at a time. This is achieved in~\cite{torii2011visual} by using the closest pair of images and then performing linear combination of them. Other works such as~\cite{brahmbhatt2018geometry} use techniques like Pose-Graph Optimization (PGO)~\cite{carlone2016planar} to take advantage of extra information available (in this case the relative poses of the ``test''). Note that these approaches differ from ours as they are used only on the query data. As such, they could be combined with the proposed method, that also considers the reference database.

\textbf{GSP:} graph signal processing~\cite{shuman2013emerging} is a mathematical framework that aims at extending harmonic analysis to irregular domains described using similarity graphs. As such, it is possible to define tools such as translations~\cite{translations}, convolutions~\cite{graphconvolutions}, filtering~\cite{filtering} and wavelets~\cite{wavelets} taking into account the complex structure of the inputs. GSP has successfully been applied to domains ranging from neuroimaging~\cite{menoret} to deep learning~\cite{graphconvolutions,smoothness,anirudh2017influential}. To our knowledge, the present work is the first usage of GSP in the context of indirect visual localization.

\section{Methodology}
\label{methodo}

In this section, we first describe the setting in which the current solution is applied and then present a formal overview of the GSP techniques as they are applied to the problem of visual localization.

\subsection{Problem Setting}

We consider the case of autonomous driving where a fleet of vehicles move around established roads in urban environments. This is a restricted setting than the more general case of localizing a freely moving tourist in a city using a mobile phone. 
Indeed the geometry of the road structure prevents significant view point variations.
In our case the change in viewpoint comes from traffic moving in different lanes along the same road. However, there might be significant viewpoint changes as vehicles can move during any season and any time of the day.

The camera mounted on the vehicle provide a stream of images, that is, we have information about the temporal adjacency of images. 
In addition, we are also provided additional information in the form of GPS location for each image.
%\CLc{Yasir I've removed the part about magnetic north because we did not integrated in this work. I believe they should give better results, but for the moment we are not using them. If needed I can try to crunch them in as quickly as possible (using them as a threshold for example)}%, and the heading of the vehicle with respect to magnetic north. 

For image representation, we assume a mapping function that maps each image to a fixed dimensional latent space, with some resilience to viewpoint and appearance changes. For the rest of the section, we use images and latent representation interchangeably to mean a lower dimensional embedding of the original image into a resilient subspace. A key asset of the latent space is that it \emph{linearizes} representations. As such, by taking the linear combination of latent representations of actual images, we usually obtain a latent representation of a natural looking (artificial) image.

\subsection{Graph Signal Processing and Graph Signals Smoothing}

In this work we consider graphs defined as tuples $G = \langle V, \mathbf{A}\rangle$, 
where $V$ is the finite set of vertices and $\mathbf{A}$ is the weighted adjacency matrix: $\mathbf{A}[\mu\nu]$ is the weight of the edge between vertices $\mu$ and $\nu$, or 0 if no such edge exists. Vertices are associated one-to-one with images, and an edge defines the similarity between two vertices.

In order to avoid irregular artifacts, we consider a normalized adjacency matrix $\mathbf{A} = \mathbf{D}^{-1}\mathbf{W}$ where $\mathbf{W}$ is the direct measure of similarity between two vertices and $\mathbf{D}$ is the degree matrix associated with $\mathbf{W}$:
 \[
     \mathbf{D}[\mu\nu] = \left\{ \begin{array}{cl}\displaystyle{\sum_{k \in V}{\mathbf{W}[\mu k]}} & \text{if } \mu = \nu\\ 0 & \text{otherwise}\end{array}\right.\;.
 \]
Note that this normalization is only well-defined if the graph has no isolated vertex, what we consider to be true in the following.

Given a graph $G = \langle V, \mathbf{A}\rangle$, consider a matrix $\mathbf{s}\in \mathbb{R}^{V\times d}$, where $d \in \mathbb{N}$. We refer to $\mathbf{s}$ as a signal in the remaining of this work, and typically we consider $\mathbf{s}$ to be composed of the concatenation of latent representations of images corresponding to vertices in $V$. As such, a row of $\mathbf{s}$ corresponds to an image in the dataset, 
whereas a column correspond to a dimension of the feature vectors representing the images. 
We define the {\em graph smoothing} $h_G(s)$ of $\mathbf{s}$ as:
\begin{equation}
h_G(\mathbf{s}) = \mathbf{A}^m\mathbf{s}.    
\label{equ:smoothin}
\end{equation}

Graph smoothing simply consists of multiplying the normalized matrix $\mathbf{A}$ of the graph with the signal. This operation can be repeated multiple times (represented by the parameter $m$). Note that smoothing can be achieved in other ways, 
for example using low-pass filters~\cite{shuman2013emerging} on the graph, 
which can be computationally expensive for large graphs. 
In this work, we focus on this particular smoothing method (\ref{equ:smoothin}) for its simplicity and performance.

Let us explain briefly why this operation has the effect of smoothing the representations in $\mathbf{s}$. First note that because $\mathbf{A}$ is symmetric and real-valued, it admits $|V|$ eigenvalues (where $|\cdot|$ denotes the cardinal). The way $\mathbf{A}$ has been normalized, all these eigenvalues are between -1 and 1. Other interesting properties include that the eigenspace associated with the eigenvalue 1 is composed of constant vectors and -1 is not an eigenvalue if the graph is not bipartite.

So, considering the graph is not bipartite, multiplying the signal by $\mathbf{A}$ has the effect of diminishing the influence of all components of the signal that are not aligned with a constant vector, while maintaining the latter. As a result, the difference between representations of neighboring vertices in the graph is reduced. This operation has the effect of smoothing the signal, in the sense that the $i$-th column of the smoothed signal is such that the difference in values between two (strongly) connected vertices is going to be smaller than that before smoothing. In the extreme case of smoothing multiple times (i.e. large $m$), this would eventually have the effect of averaging all representations in connected components of the graph.

\begin{figure}[h!]
  \begin{center}
    \begin{tikzpicture}[thick]      
    \begin{scope}[scale=1.4]
    \begin{scope}[xshift=0cm]
        \node[draw, inner sep = 2pt, circle](a) at (-0.5,0.5) {};
        \node[draw, inner sep = 2pt, circle](b) at (-0.25,0) {};
        \node[draw, inner sep = 2pt, circle](c) at (0.25,0.25) {};
        \node[draw, inner sep = 2pt, circle](d) at (0.75,0.25) {};
        \node[draw, inner sep = 2pt, circle](e) at (1.0,0.6) {};
        \node[draw, inner sep = 2pt, circle](f) at (1.25,0.85) {};
        \node[draw, inner sep = 2pt, circle](g) at (1.5,0.) {};
                \path[]
        (-0.5, 0.5) edge[blue] (-0.5, 1.42)
        (-0.25, 0.) edge[blue] (-0.25, 1.02)
        (0.25, 0.25) edge[red] (0.25, 0.15)
        (0.75, 0.25) edge[blue] (0.75, 0.4)
        (1.0, 0.5) edge[red] (1.0, -0.6)
        (1.25, 0.75) edge[red] (1.25, 0.)
        (1.5, 0.) edge[red] (1.5, -1.2);
        \node[fill, circle, inner sep = 1pt, blue] at (-0.5, 1.42) {};
        \node[fill, circle, inner sep = 1pt, blue] at (-0.25, 1.02) {};
        \node[fill, circle, inner sep = 1pt, red] at (0.25, 0.15) {};
        \node[fill, circle, inner sep = 1pt, blue] at (0.75, 0.4) {};
        \node[fill, circle, inner sep = 1pt, red] at (1.0, -0.6) {};
        \node[fill, circle, inner sep = 1pt, red] at (1.25, 0.) {};
        \node[fill, circle, inner sep = 1pt, red] at (1.5, -1.2) {};
        
      \end{scope}
      \node at (-0.5,-0.5) {};
      \path[black!20!white]
      (a) edge (c)
      (b) edge (c)
      (c) edge (d)
      (d) edge (e)
      (d) edge (g)
      (e) edge (f);
      \node at (2.5, 0.5) {\Huge{$h_G(\mathbf{s}) \atop \Rightarrow$}};
      \begin{scope}[xshift=4cm]
        \node[draw, inner sep = 2pt, circle](a) at (-0.5,0.5) {};
        \node[draw, inner sep = 2pt, circle](b) at (-0.25,0) {};
        \node[draw, inner sep = 2pt, circle](c) at (0.25,0.25) {};
        \node[draw, inner sep = 2pt, circle](d) at (0.75,0.25) {};
        \node[draw, inner sep = 2pt, circle](e) at (1.0,0.6) {};
        \node[draw, inner sep = 2pt, circle](f) at (1.25,0.85) {};
        \node[draw, inner sep = 2pt, circle](g) at (1.5,0.) {};
                \path[]
        (-0.5, 0.5) edge[blue] (-0.5, 1.0)
        (-0.25, 0.) edge[blue] (-0.25, 0.8)
        (0.25, 0.25) edge[blue] (0.25, 0.4)
        (0.75, 0.25) edge[red] (0.75, 0.1)
        (1.0, 0.5) edge[red] (1.0, 0.3)
        (1.25, 0.75) edge[red] (1.25, 0.2)
        (1.5, 0.) edge[red] (1.5, -0.8);
        \node[fill, circle, inner sep = 1pt, blue] at (-0.5, 1.0) {};
        \node[fill, circle, inner sep = 1pt, blue] at (-0.25,0.8) {};
        \node[fill, circle, inner sep = 1pt, blue] at (0.25, 0.4) {};
        \node[fill, circle, inner sep = 1pt, red] at (0.75, 0.1) {};
        \node[fill, circle, inner sep = 1pt, red] at (1.0, 0.3) {};
        \node[fill, circle, inner sep = 1pt, red] at (1.25, 0.2) {};
        \node[fill, circle, inner sep = 1pt, red] at (1.5, -0.8) {};

      \end{scope}
      \node at (-0.5,-0.5) {};
      \path[black!20!white]
      (a) edge (c)
      (b) edge (c)
      (c) edge (d)
      (d) edge (e)
      (d) edge (g)
      (e) edge (f);
      
    \end{scope}
    \end{tikzpicture}
    \label{graph_smoothing}
    \caption{Illustrative example of the graph smoothing operation. The signal is represented by the blue(positive) and red(negative) bars.}
    \end{center}
\end{figure}
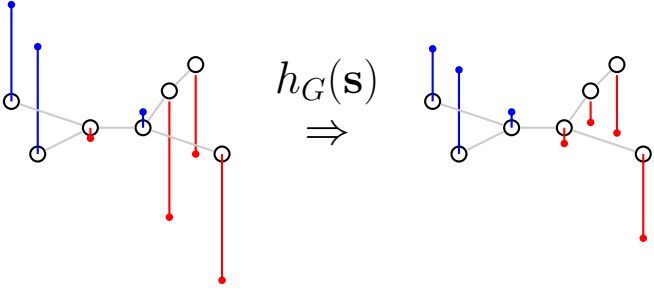

In brief, graph smoothing has the effect of smoothing the signal values, taking into account strongly connected vertices in the graph. As a result, outliers are smoothed using similar images in the graph. In this work, we consider the vertices to be either the reference database or the query database. In both cases, the goal is to use graph smoothing to reduce the noise in the latent representations. This is illustrated in Figure~\ref{graph_smoothing}, where we consider a unidimensional signal represented using blue (for positive values) and red (for negative values) bars. Before smoothing (on the left), neighboring vertices can have large variations in their signal values. After smoothing (on the right), these variations are lowered. Note that the parameter $m$ in Equation~(\ref{equ:smoothin}) controls the intensiveness of smoothing: when $m$ is small (i.e. almost 0), $\mathbf{A}^m$ becomes close to the identity matrix and the smoothing has almost no effect. When $m$ is large (i.e. $m\gg 1$), $\mathbf{A}^m$ becomes an averaging matrix on each of its connected components.

\subsection{Graph definition}

In order to make the graph smoothing improve the accuracy of VBL, we need to make sure that the edges of the graph are well chosen to reflect the similarity between two images represented as vertices, as our main goal is to exploit extra information available at the database. In this work, we consider three different sources:

\begin{itemize}
    \item Metric distance (\texttt{dist}): the distance measured by the GPS coordinates between vertices $\mu$ and $\nu$;
    \item Sequence (\texttt{seq}): the distance in time acquisition between two images (acquired as frames in videos);
    \item Latent similarity (\texttt{latent\_sim}): the cosine similarity between latent representations.
\end{itemize}

The matrix $\mathbf{W}$ can therefore be derived from the three sources as: $$\mathbf{W} = \mathbf{W}_{\texttt{dist}} + \mathbf{W}_{\texttt{seq}} + \mathbf{W}_{\texttt{latent\_sim}}.$$

\subsubsection{Metric distance}

In order to transform the metric distance into a similarity, we use an exponential kernel. This is parametrized by a scalar $\alpha$ that controls the sharpness of the exponential and a threshold parameter $max_{distance}$ that cuts edges between distant vertices:
$$\mathbf{W}_{\texttt{dist}}[\mu \nu] = \left\{ \begin{array}{ll} e^{\alpha dist_{\mu,\nu}} & \text{if } dist_{\mu,\nu} < max_{distance}\\ 0 & \text{otherwise}\end{array}\right..$$

Note that the choice of an exponential kernel may seem arbitrary, but is often used in the area of Graph Signal Processing~\cite{shuman2013emerging}.

\subsubsection{Sequence}

To exploit the information of time acquisition of frames, we use the function $seq(k, \mu, \nu)$ which returns 1 if the frame distance between $\mu$ and $\nu$ is exactly $k$ and 0 otherwise. We then build a matrix $\mathbf{W}_{\texttt{seq}}$ parametrized by scalars $\beta_k$ and $k_{max}$:

$$\mathbf{W}_{\texttt{seq}}[\mu \nu] = \sum_{k=1}^{k_{max}} \beta_{k} seq(k,\mu, \nu).$$

\subsubsection{Latent similarity}

Finally, we define a matrix $\mathbf{W}_{\texttt{latent\_sim}}$ for the latent representations cosine similarity. This is parametrized by a scalar $\gamma$ that controls the importance of the latent similarity. We only compute this similarity if either the distance similarity or the sequence similarity is nonzero:

$$\mathbf{W}_{\texttt{latent\_sim}}[\mu \nu] = \left\{ \begin{array}{ll} \gamma sim(\mu,\nu) & \text{if } W_{dist}[\mu \nu] > 0 \\ & \text{or } W_{seq}[\mu \nu] > 0,\\ 0 & \text{otherwise}\end{array}\right..$$

where $sim$ is the latent similarity function. In this work we use the cosine similarity, but any similarity function could be used.

%$$\mathbf{W}_{\texttt{latent\_sim}}[\mu \nu] = \left\{ \begin{array}{ll} \gamma cos_{sim}(\mu,\nu) & \text{if } dist_{\mu,\nu} < max_{distance} \atop or \sum_{k=1}^{k_{max}} seq(k,\mu, \nu) > 0\\ 0 & \text{otherwise}\end{array}\right..$$

\subsection{NetVLAD}

For image representation in a latent space, we use features from a pretrained NetVLAD~\cite{arandjelovic2016netvlad}, trained on the Pittsburgh dataset~\cite{Torii-PAMI2015,Torii-CVPR2013}. 
The model is available online at~\cite{GithubNetVLAD}.
NetVLAD is specifically trained to cater for viewpoint and appearance changes.
It maps an image to $32768$ dimensional deeply learnt representation, 
which we then compressed to $4096$ dimensions using PCA (trained on the support database) and then finally whitened~\cite{pcawhiten}. We follow the same image preprocessing from the training of the model, where images are first resized so that the smaller part has a size of 256 pixels, then we perform a center-crop of 224, and finally perform standardization. 

\section{Experiments}
\label{experiments}

\subsection{Dataset generation}

\begin{figure}
    \centering
    \includegraphics[width=\columnwidth]{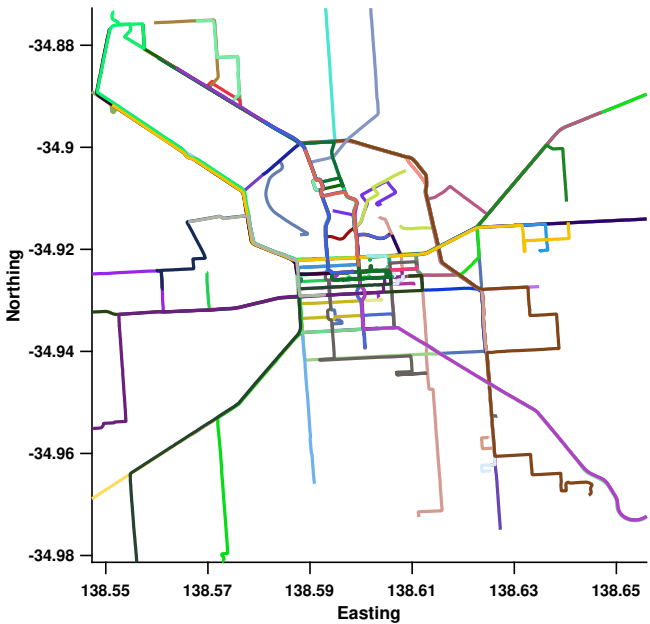}
    \caption{GPS Tracks of image sequence collected around Adelaide CBD from Mapillary.}
    \label{fig:CBD}
\end{figure}

\begin{figure}
    \centering
    \includegraphics[width=\columnwidth]{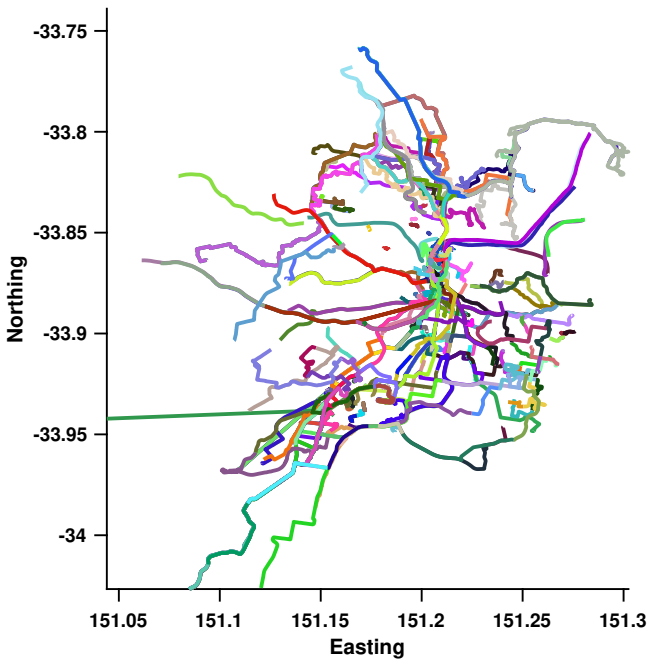}
    \caption{GPS Tracks of image sequence collected around Sydney from Mapillary.}
    \label{fig:Sydney}

\end{figure}

In order to verify the effectiveness our method in the setting of autonomous driving,
we need a dataset that is collected from roads and is large enough to demonstrate appearance changes and limited viewpoint changes due to road structure.
We collect images from  Mapillary API\footnote{\url{https://www.mapillary.com/developer/api-documentation/}},
which contains publicy sourced data over time for major roads.
To show the generalization ability of the proposed work, we collect road imagery from two Australian cities. 
The first covers the Central Business District (CBD) area of Adelaide, Australia and spans an area of roughly 10km$^2$. Since the data is publicly sourced, there is a lot of viewpoint, illumination and dynamic changes (cars, pedestrian, etc).
The second set is collected around the Greater Sydney region and covers an area of around 200km$^2$. We note that the data collected for the Greater Sydney region contains some sequences that were generated using different equipment (panoramic cameras) or different positioning (camera pointed to a vehicule window instead of the windshield) from the ones used during the training of the NetVLAD network, which combined with the total area of the support database creates a much more challenging problem. 
In addition to imagery, the collected data provides sequence information and GPS. % and heading of the vehicle with respect to the magnetic north. 
The GPS tracks for the collected data are shown in Figures~\ref{fig:CBD} and ~\ref{fig:Sydney}.

In the rest of the experiment section,
we use the terminology of indirect visual localization, that is, 
support database refers to the reference database, 
validation and test queries refer to query inputs. 

To split the Adelaide dataset in support/validation/test we randomly choose 4 sequences for validation and 5 sequences for testing.
For the Sydney database, we choose 5 sequences that could be retrieved with reasonable performance using our pre-trained NetVLAD (named easy query) 
and 5 sequences at random (hard query). 
Using GPS as ground truth, we remove all examples from the query sets that are further than 25m from the support dataset (i.e. there are no examples in the support set in a 25m radius from them)\footnote{The dataset and code for reproducing the results will be made public on acceptance of the work.}.
The statistics for each dataset are summarized in table~\ref{dataset-summary}. 

\begin{table}[]
\centering
\caption{Summary of the datasets used in this work}
\begin{tabular}{|c|cc|}
\hline
City             & \multicolumn{2}{c|}{Adelaide} \\ \hline
                 & \# Sequences     & \# Images    \\ \hline
Support Database & 44              & 24263       \\ 
Validation Query & 4               & 2141        \\ 
Test Query       & 5               & 1481        \\ \hline
                 & \multicolumn{2}{c|}{Sydney} \\ \hline
                 & \# Sequences     & \# Images    \\ \hline
Support Database & 284              & 117860       \\ 
Easy Query       & 5               & 1915      \\ 
Hard Query       & 5        & 2285   \\ \hline
\end{tabular}
\label{dataset-summary}
\end{table}

\subsection{Parameter definition}

For all the results in the subsequent sections we use the same parameters, which were obtained using a grid search and keeping the best score on the Adelaide validation query. We use the Adelaide test query to ensure that the parameters are not overfitted to the validation query. Also note that we use the same parameters for all cities to further validate the fact that we do not need additional training/parameter search for each new city. The parameters are $\alpha = 0.25, \beta_1 = 0.75, \beta_2 = 0.0625, \beta_3 = 0.0625, k_{max} = 3, \gamma = 0.33, m=2$.

\subsection{Results}

We test the graph smoothing effect in three different cases. First the extra data is available only for the support, second it is available only for the query and finally it is available in both cases. In each case we report two metrics, the median localization error over all the queries and the percentage of localizations that have less than 25m error.

First we perform the tests on the Adelaide dataset and present the results in table~\ref{results_adelaide}. The graph smoothing operation was able to increase performance, even when applied only on the query database, and as expected, adding the graph smoothing during both query and support gave the best results. Recall that the parameters were defined based on the validation query, under the case where the extra data is available only for the support database.

\begin{table}[h]
\centering
\caption{Results under different graph smoothing conditions for the Mapiliary Adelaide dataset. GS means graph smoothing}
\begin{tabular}{|c|c|c|c|c|}
\hline
Measure                                & None & GS Support & GS Query & GS S+Q \\ \hline
\multicolumn{5}{|c|}{Validation}                           \\ \hline
acc $<$ 25m & 66.84\%           & 74.64\%                & 70.06\%              & \textbf{79.03\%}                \\ 
median distance      & 8.76m            & \textbf{7.29m}                 & 13.02m              & 9.17m                 \\ \hline
\multicolumn{5}{|c|}{Test}                           \\ \hline
acc $<$ 25m       & 44.63\%           & 50.03\%                & 46.39\%              & \textbf{51.32\%}                \\ 
median distance            & 110.66m          & 24.08m                & 41.84m              & \textbf{22.81m}                \\ \hline
\end{tabular}
\label{results_adelaide}
\end{table}

Second we validate that the operation can be used on other cities and that we do not need to perform an additional grid search for the new data. The results are presented in Table~\ref{results_sydney}. As expected the graph smoothing operation allowed us to get better performance in both median distance and accuracy, while using the parameters optimized for the Adelaide dataset. This is inline with our goal that is to have an operation that we do not have to retrain or re-validate parameters for a new dataset. We note that the performance of the hard query set is not inline with a good retrieval system (several kilometers from the correct point), but it is included to show that our method allows us to increase the performance both when the NetVLAD features are already very good for the task and when they are very bad. 

\begin{table}[h]
\centering
\caption{Results under different graph smoothing conditions for the Mapiliary Sydney dataset. GS means graph smoothing}
\begin{tabular}{|c|c|c|c|c|}
\hline
Measure                                & None & GS Support & GS Query & GS S+Q \\ \hline
\multicolumn{5}{|c|}{Easy}                           \\ \hline
acc $<$ 25m & 49.45\%           & 56.16\%                & 55.93\%              & \textbf{64.21\%}                \\ 
median distance      & 28.25m            & 13.48m                 & 18.41m              & \textbf{12.13m}                 \\ \hline
\multicolumn{5}{|c|}{Hard}                           \\ \hline
acc $<$ 25m       & 13.87\%           & 17.33\%                & 16.67\%              & \textbf{24.07\%}                \\ 
median distance            & 4000km          & 3373m                & 3149m              & \textbf{2151m}                \\ \hline
\end{tabular}
\label{results_sydney}
\end{table}

\subsection{Ablation studies}

To verify that each part of the graph is important, we perform ablation studies using the Adelaide test query. The results are presented in Table~\ref{ablation}. 
The table shows that different sources of information are important, with each one adding to increase in performance. 
Metric distance and sequence being the most important features and latent similarity being more of a complementary feature (this is expected, as it is being thresholded by the other two features). This is encouraging since in the absence of any other external information (GPS, etc), one can rely on the sequential nature of data collection to get a boost in localization performance. This information is readily available in a robotics setting. 

\begin{table}[h]
\centering
\caption{Ablation study on the Mapiliary Adelaide test query.}
\begin{tabular}{|c|c|c|c|c|}
\hline
$\mathbf{W}_{\texttt{dist}}$ & $\mathbf{W}_{\texttt{seq}}$ & $\mathbf{W}_{\texttt{latent\_sim}}$ & median distance & acc $<$ 25m \\ \hline
     &     &     & 110.66m          & 44.63\%            \\ \hline
X    &     &     & 46.10m           & 47.26\%            \\ \hline
     & X   &     & 39.11m           & 47.53\%            \\ \hline
X    &     & X   & 42.92m           & 47.60\%            \\ \hline
X    & X   &     & 24.75m           & 50.03\%            \\ \hline
     & X   & X   & 37.39m           & 47.47\%            \\ \hline
X    & X   & X   & \textbf{24.08m}  & \textbf{50.30\%}   \\ \hline
\end{tabular}
\label{ablation}
\end{table}

\begin{figure}[ht]
 \begin{center}
   \begin{tikzpicture}
       \begin{axis}[
           xlabel=$m$,
           ylabel=acc $<$ 25,
           xmin=0]
         \addplot table {ablation_m.txt};
       \end{axis}
   \end{tikzpicture}
    \caption{Effect of the parameter $m$ on the retrieval accuracy under 25m for the Adelaide test query.}
    \label{ablation_m}
 \end{center}
\end{figure}
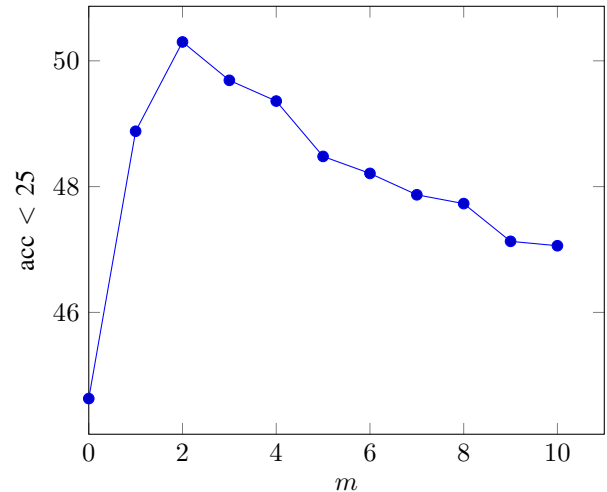

In the next experiment, we demonstrate the effect of successive smoothing. 
This is achieved by applying smoothing operation $m$ times. 
Theoretically, this should help increase the performance until it hits a ceiling and then it should start to slowly decrease (as it enforces connected examples of the database to be too similar to each other). 
The results are presented in 
Fig.~\ref{ablation_m}. 
As can be seen, there is a clear pattern of increased performance until $m=2$ after which the performance starts to degrade. It should be noted that even for $m=10$ the graph smoothing operation still performs better than the baseline ($m=0$).

\section{Conclusion}
\label{conclusion}

This work showed that using techniques from Graph Signal Processing, the performance of indirect visual based localization can be improved by incorporating additional available information. 
This additional information acts on the latent representation by making it smoother on a graph designed using all available information, leading to a boost in localization.
One of the encouraging observation of the work is that this additional information can take the form of a simple temporal relationship between surrounding images acquired in a sequence, and still lead to a significant increase in performance.

%As we described in the introduction, adding graph smoothing can be an interesting way to increase the performance of deep representations for indirect visual localization. 
In future work, we would like to use the graph during the localization inference, to add temporal consistency to the position inference and also to train the smoothing operation in an end-to-end fashion.

\bibliographystyle{IEEEtran}
\bibliography{IEEEabrv,root}

\addtolength{\textheight}{-12cm}   % This command serves to balance the column lengths

%%%%%%%%%%%%%%%%%%%%%%%%%%%%%%%%%%%%%%%%%%%%%%%%%%%%%%%%%%%%%%%%%%%%%%%%%%%%%%%%

%%%%%%%%%%%%%%%%%%%%%%%%%%%%%%%%%%%%%%%%%%%%%%%%%%%%%%%%%%%%%%%%%%%%%%%%%%%%%%%%

%%%%%%%%%%%%%%%%%%%%%%%%%%%%%%%%%%%%%%%%%%%%%%%%%%%%%%%%%%%%%%%%%%%%%%%%%%%%%%%%
%\section*{APPENDIX}

%\section*{ACKNOWLEDGMENT}

%%%%%%%%%%%%%%%%%%%%%%%%%%%%%%%%%%%%%%%%%%%%%%%%%%%%%%%%%%%%%%%%%%%%%%%%%%%%%%%%

\end{document}